\documentclass{article}
\usepackage[preprint]{neurips_2023}
\usepackage{amssymb}
\usepackage{amsmath}
\usepackage{lastpage}
\usepackage{dsfont}
\usepackage{float}
\usepackage{caption}
\usepackage{xcolor}
\usepackage{framed}
\usepackage{tcolorbox}
\usepackage{booktabs}
\usepackage{multirow}
\usepackage{subcaption}
\usepackage{rotating}
\usepackage{booktabs}
\usepackage{tabularx}
\usepackage{pdfpages}
\usepackage{color}
\usepackage[toc,page]{appendix}
\usepackage{gensymb}
\usepackage{epstopdf, epsfig}
\usepackage{hyperref}
\usepackage{pgfplots}
\pgfplotsset{compat=newest}
\usepackage{multicol}  
\usepackage{bm}	
\usepackage{pdflscape}
\usepackage[T1]{fontenc}
\usepackage{ae,aecompl}
\usepackage{pdflscape}

\author{
  Giorgi Kokaia\\
  Ingka (IKEA) \\
  \texttt{giorgi.kokaia@ingka.ikea.com} \\
  \And
  Pratyush Sinha\\
  Ingka (IKEA) \\
  \texttt{pratyush.sinha@ingka.ikea.com} \\
  \And
  Yutong Jiang \\
  KTH Royal Institute of Technology \\
  \texttt{yutongj2@gmail.com} \\
  \And
  Nozha Boujemaa \\
  Ingka (IKEA) \\
  \texttt{nozha.boujemaa@ingka.ikea.com}
}

\title{Writing your own book: A method for going from closed to open book QA to improve robustness and performance of smaller LLMs}
\date{\today}

\begin{document}
\label{firstpage}
\maketitle

\begin{abstract}
We introduce two novel methods, Tree-Search and Self-contextualizing QA, designed to enhance the performance of large language models (LLMs) in question-answering tasks. Tree-Search is a sampling technique specifically created to extract diverse information from an LLM for a given prompt. Self-contextualizing QA leverages Tree-Search to enable the model to create its own context using a wide range of  information relevant to the prompt, evaluate it explicitly and return a open book answer to the initial prompt . We demonstrate that the quality of generated answers improves according to various metrics, including accuracy, informativeness, coherence, and consistency, as evaluated by GPT3.5(text-davinci-003). Furthermore, we show that our methods result in increased robustness and that performance is positively correlated with tree size, benefiting both answer quality and robustness. Finally, we discuss other promising applications of Tree-Search, highlighting its potential to enhance a broad range of tasks beyond question-answering.

\noindent We also discuss several areas for future work, including refining the Tree-Search and Self-Contextualizing QA methods, improving the coherence of the generated context, and investigating the impact of bootstrapping on model robustness\end{abstract}

\section{Introduction}
The most notable breakthrough in artificial intelligence research over the past few years has been the significant progress in natural language processing (NLP) driven by large language models (LLMs). These transformer-based neural networks \cite{Vawani2017} are trained on enormous corpora of web-text data, utilizing a self-supervised objective that involves predicting the next word in a given partial sentence. As a result, LLMs, such as BERT \cite{Devilin2019}, GPT-3 \cite{Brown2020}, and GPT-4, have demonstrated exceptional performance across a wide array of NLP tasks, including machine translation, sentiment analysis, text summarization, and question-answering (QA) \citep{yang2019xlnet, Sparks2023}.

Despite the remarkable achievements of LLMs, they still face challenges in robustness, context understanding, and generalization, particularly in question-answering tasks under closed book and open book settings \citep{Chen2017}. Closed book QA systems derive answers solely based on the internal knowledge gained during the pre-training phase of the model, while open book QA systems leverage external information sources, such as knowledge bases or documents, to provide more accurate and contextually relevant responses.

In this study, we introduce two novel methodologies: Tree-Search and Self-Contextualizing QA. Tree-Search is a new decoding strategy designed to extract a diverse range of information from a given model, enabling the generation of richer context for question-answering tasks. Self-Contextualizing QA refers to the process of transforming closed book QA into open book QA by creating context from the model's own outputs. By combining these two methodologies, we aim to enhance the performance and robustness of large-scale language models in QA tasks.
\section{Closed book vs Open book}\label{sec:robust}
There is a large number of different tasks on which LLMs are trained, all of which help build their capabilities and knowledge (see e.g.~\citealp{raffel2020} for a good breakdown). We consider two of them, closed book QA and open book QA. 

Closed book QA refers to a setting where the model generates answers based on its internal knowledge acquired during the pre-training process, without access to external information (e.g.~\citealp{Chen2017}). LLMs such as GPT-3~\citealp{Brown2020}, have shown remarkable performance in closed book QA tasks due to their ability to store and recall vast amounts of knowledge.

Open book QA involves providing the model with access to external information, such as documents or databases, which it can utilize to generate more accurate and up-to-date answers (e.g.~\citealp{Lewis2020}). This extra information is generally referred to as \textit{context}.

There are of course some advantages and disadvantages to each approach. Closed book QA will be faster and require less computational resources as it accesses knowledge stored within its weights. This also means that is limited to the this knowledge, which can result in outdated or incomplete answers~\citep{Roberts2020} making open book QA more reliable for the types of tasks that require up to date, domain specific knowledge~\citep{Thorne2018}. Additionally, given the required time as well as the computational and environmental costs associated with training the largest LLMs (as discussed in e.g.~\citealp{Touvron2023}) it becomes completely unfeasible to constantly retrain them to keep them up to date making some form of open book QA a must.

A crucial aspect distinguishing open book QA from closed book QA is the increased robustness exhibited by the former. In this context, robustness refers to the ability of the model to generate answers that are less sensitive to small changes in the input prompt. This issue has been identified even in the largest models, such as GPT-4 \citep{Sparks2023}.

In yet to be published work by \textit{Jiang et al., (2023)}, an experiment was conducted where a model (T5) was asked the same question with and without a provided context. The results indicated that the open book QA exhibited a larger difference in probabilities between the top two tokens in the first position, signifying greater confidence in the prediction. This increased certainty is crucial because small variations in the input prompt can lead to different probabilities for the top predicted tokens. Given the auto-regressive nature of language model predictions, this shift in probabilities can cause a cascading change in the entire prediction. Open book QA's enhanced stability in token predictions makes it a more robust approach, which is a key reason for why this study focuses on developing a method to transform closed book QA into open book QA.

By transforming closed book QA to open book QA, we aim to not only improve the quality of generated answers but also enhance the overall robustness of the model's behavior. This increased robustness should result in more reliable and consistent predictions, which are less sensitive to minor variations in the input prompts. We reproduce and extend the experiment by \textit{Jiang et al., (2023)} in our work to further demonstrate the benefits of this approach.

\section{Tree-Search Method}\label{sec:tree}
Tree-Search is a sampling method which aims to extract the most varied information possible from a LLM when given a specific prompt. This approach is particularly useful when seeking diverse responses from the model in order to explore a broader range of solutions or insights. The Tree-Search method can be broken down into three main steps (identifying high entropy positions, creating branches, and iterating the process), which are described in detail below.
\subsubsection{Identifying High Entropy Positions}
The first step in the Tree-Search method is to identify high entropy positions within the model's decoded output. High entropy positions represent points where the model has low confidence in its predictions, making them ideal for branching and exploration. We propose two ways to identify these positions:
\begin{enumerate}
    \item \textbf{Relative Probability Threshold:} Calculate the relative probabilities between the top tokens at each position. If the relative probability falls below a pre-defined threshold (near unity), it indicates that the model has low confidence at that position, and hence, it is a high entropy position.
    \item \textbf{Probability Cut-off:} Set a probability cut-off at a reasonably low value (e.g., ~0.01). Count the number of remaining tokens at each position after applying this cut-off. If the number of tokens exceeds a certain threshold, it indicates that the position is a high entropy position.
\end{enumerate}
\subsubsection{Creating Branches}\label{sec:branch}
Once high entropy positions are identified, the next step is to create branches in the decoding process. This can be done in one of two ways:
\begin{enumerate}
    \item \textbf{Non-Greedy Token Selection:} For each high entropy position, select any token below the threshold instead of the highest probability one. This encourages exploration of less probable but potentially interesting and relevant outputs.
    \item \textbf{Random Token Selection:} Alternatively, for each high entropy position, randomly select a token above the probability cut-off. This approach promotes diversity in the generated responses.
\end{enumerate}
\subsubsection{Iterating the Process}
The final step of the Tree-Search method involves repeating the branching process. Continue to create branches either until no more branches can be formed (as the criteria for high entropy positions are no longer met) or until a desired depth is reached. The depth represents the number of times each sequence has been branched and can be adjusted to control the extent of exploration. We call a complete output (i.e. an output that includes </eos>) a leaf.

\subsection{Tree-Search versus Traditional Beam Search}
\begin{figure}
    \centering
    \includegraphics[width=0.75\columnwidth]{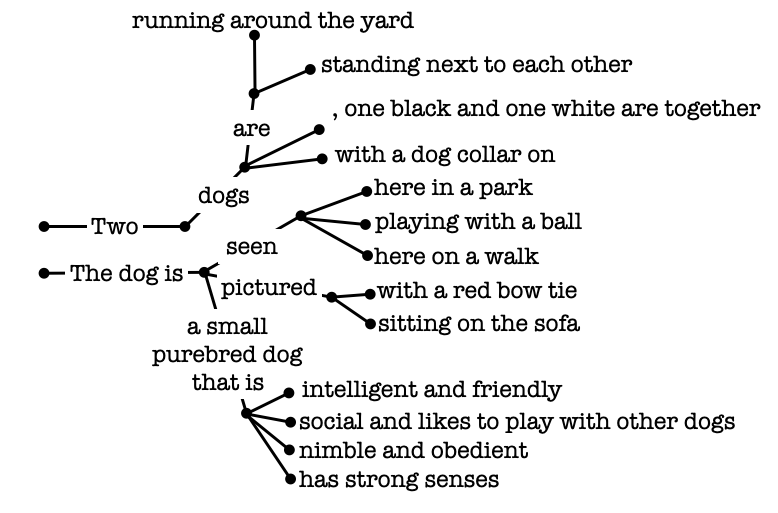}
    \caption{The tree generated from the prompt "Describe the features of a dog" using tree search with random token selection as described in section~\ref{sec:branch}. It should be noted that in this particular case the tree branched at the very first token, giving it the appearance of two separate outputs, although this is not the case. }
    \label{fig:tree}
\end{figure}
Beam search is a widely used technique for generating sequences in AI models. It works by expanding the search space in a breadth-first manner, maintaining a fixed number of top candidate sequences (called ”beams”) at each step. Beam search aims to strike a balance between computational efficiency and the quality of generated sequences. However, it tends to produce less diverse outputs, as it follows a more focused search strategy, retaining only the most likely sequences at each step. Which is a fundamentally different strategy from what we are proposing in this study.

Contrarily, the proposed Tree-Search approach aims for diverse and exploratory outputs by targeting high entropy positions during the model's decoding process. By branching at these positions and adopting non-greedy or random token selection, Tree-Search explores less probable but potentially intriguing solutions, thereby generating a wider range of responses.

The distinguishing factors between Tree-Search and beam search include the output diversity, with Tree-Search yielding more varied results; exploration versus exploitation, where Tree-Search fosters search space exploration while beam search exploits the model's most probable predictions; and customizability, where Tree-Search provides more user control over exploration depth and output diversity, whereas beam search primarily focuses on maintaining a fixed number of top sequences.

We illustrate the differences by showing the resulting tree from a very simple prompt, "describe the features of a dog", in figure~\ref{fig:tree}. Whilst the answers are not what one might expect, they do answer the prompt and they do so with a rather large variety. In fact, the greedy answer which is "The dog is a member of the Canidae family" is both a worse answer to the prompt and incidentally does not even appear in the tree. We do not show the comparison with beam search in the figure as performing a beam-search with 100 beams and picking the top 10 beams gives essentially the same greedy answers with minor variations to it.

\section{Setup of the experiment}\label{sec:experiment}

Throughout this study we conduct all the experiments using the model T0\_3B~\citep{Sanh2022}. In the main experiment we apply the Tree-Search to create context that transforms closed book QA prompts into open book QA prompts. The goal of this experiment is to assess whether providing context generated by Tree-Search leads to better answers and more robust behavior. The experiment is as follows:
\begin{figure}[H]
    \centering
    \includegraphics[width=\textwidth]{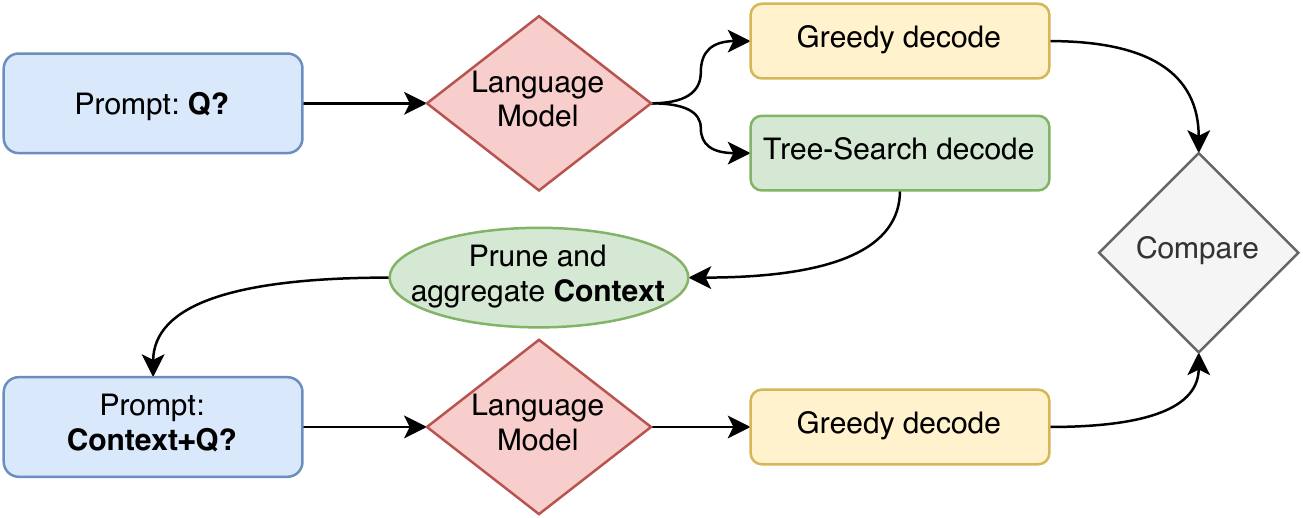}
    \caption{The flowchart shows illustrates the process of going from closed book to open book QA described in section~\ref{sec:experiment}.}
    \label{fig:flowchart}
\end{figure}
\noindent Our process begins with the assembly of a QA dataset composed of general knowledge, open-ended questions such as "What caused the French Revolution?" We initially prompt the model with these questions, storing the responses as a baseline for comparison with answers obtained using Tree-Search and the subsequent open book QA approach. Next, we apply Tree-Search to the model using the same dataset, yielding a variety of potential answers to each question. We then take each Tree-Search output, prune all the duplicate text and concatenate all unique outputs. This becomes the context for the model that takes it from closed to open book QA.

With this context, the model is prompted again, this time following an open book QA approach that utilizes the provided context to generate answers. These responses are stored for further analysis and comparison with the initial closed book QA outputs. Lastly, we evaluate both sets of responses using GPT3.5(\textsc{text-davinci-003}), a model demonstrated to perform on par with humans in text annotation~\citep{Huang2023}. The entire process is illustrated by the flowchart in figure~\ref{fig:flowchart}
\section{Results}\label{sec:result}
We have put together a dataset of 1475 open ended general knowledge questions. We have applied the process described in the previous section to this dataset, where we build the tree using a relative probability threshold of 1.4 and doing an exhaustive Tree-Search to a depth of 3. We do however set the maximum number of branches to 20 for computational efficiancy. Once context is created we then prompt the model according to this very simple template that we arrived at following some experimentation. 
\definecolor{shadecolor}{gray}{0.9}
\begin{tcolorbox}
\noindent \textbf{Context:} \verb|{{context}}| \\
\noindent \textbf{Question:} \verb|{{question}}| \\
\noindent \textbf{Answer:}
\end{tcolorbox}

We then compare the original answer with the new one using GPT3.5(\textsc{text-davinci-003}) in four different ways 1) which answer is the most informative, 2) which is the most accurate 3) is the most consistent and 4) is the most consistent. In the prompt to \textsc{text-davinci-003} we provide the question, the original as well as the new answer and then ask it to evaluate which answer best, given the way in which we are comparing them, or if they are very similar. In an effort to minimise bias we then ask the opposite question (i.e. "which answer is the least informative"), we set the true value to be the average of these and then we bootstrap the outputs in order to get an estimate of the uncertainty. The results of this experiment can be found in table~\ref{tab:results}.

\noindent Below, two typical types of outputs are shown to two different prompts. In the first one we see that the open book provides a more detailed and informative answer as we would expect it to do. In the second one the open book also provides a more informative answer, however there we see that some errors are propagated from the simplistic way in which the context is put together. 
\definecolor{my-blue}{cmyk}{0.80, 0.13, 0.14, 0.04, 1.00}
\begin{tcolorbox}[colback=my-blue]
\noindent \textbf{Question:} \textit{What makes a work of literature timeless?}?
 \\
\noindent \textbf{Closed book answer:} It is a work of art that has stood the test of time. \\
\noindent \textbf{Open book answer:} The best works of literature are timeless. They have stood the test of time and are still relevant today, despite being written in different times or places
\end{tcolorbox}
\begin{tcolorbox}[colback=my-blue]
\noindent \textbf{Question:} \textit{What were the major accomplishments of Martin Luther King Jr.}?
 \\
\noindent \textbf{Closed book answer:} Martin Luther King Jr. was an American civil rights leader and activist. \\
\noindent \textbf{Open book answer:} Nobel Peace Prize winner who is widely regarded as one of the greatest leaders in American history, influencing many people around him to work for social justice issues such at voting rights or equal rights for women who is widely regarded as one of the greatest leaders in American history, particularly for his efforts to end segregation throughout the United States during the 1950s through the 1960'Sated in 1968 by the Ku Klux Klang. Tennessee at the age of 50 after a campaign of nonviolent protest against segregation
\end{tcolorbox}
\begin{table}[H]
\centering
      \caption{The figure shows the evaluation of closed book as well as the open book answers by GPT3.5(\textsc{text-davinci-003}). We evaluate the original answers as well as the new answers in four different ways; how much information is in it, the quality of it, its accuracy and its consistency. The evaluation is done on a total of 1475 questions.}
\begin{tabular}{lccc}

    \centering      
    Metric & Closed Book[$\%$] & Open Book[$\%$] & Same/Similar[$\%$] \\
    \hline
    \addlinespace
    Informative & $8.4^{+1.0}_{-1.4}$ & $53.3^{+21.0}_{-17.8}$ & $38.3^{+13.4}_{-11.7}$ \\
    \addlinespace
    Accuracy    & $12.5^{+1.3}_{-1.8}$ & $31.2^{+8.2}_{-7.1}$ & $56.3^{+17.8}_{-16.1}$ \\
    \addlinespace
    Coherent    & $11.9^{+1.5}_{-1.1}$ & $29.6^{+8.7}_{-7.4}$ & $58.5^{+32.7}_{-29.9}$ \\
    \addlinespace
    Consistent  & $14.1^{+3.2}_{-2.7}$ & $26.4^{+7.4}_{-7.8}$ & $59.5^{+16.7}_{-14.2}$ \\
    \addlinespace
    \hline
    \label{tab:results}
\end{tabular}
\end{table}

\section{Discussion}
\subsection{Testing robustness with our methodology}
\begin{figure}[H]
    \centering
    \includegraphics[width=0.6\columnwidth]{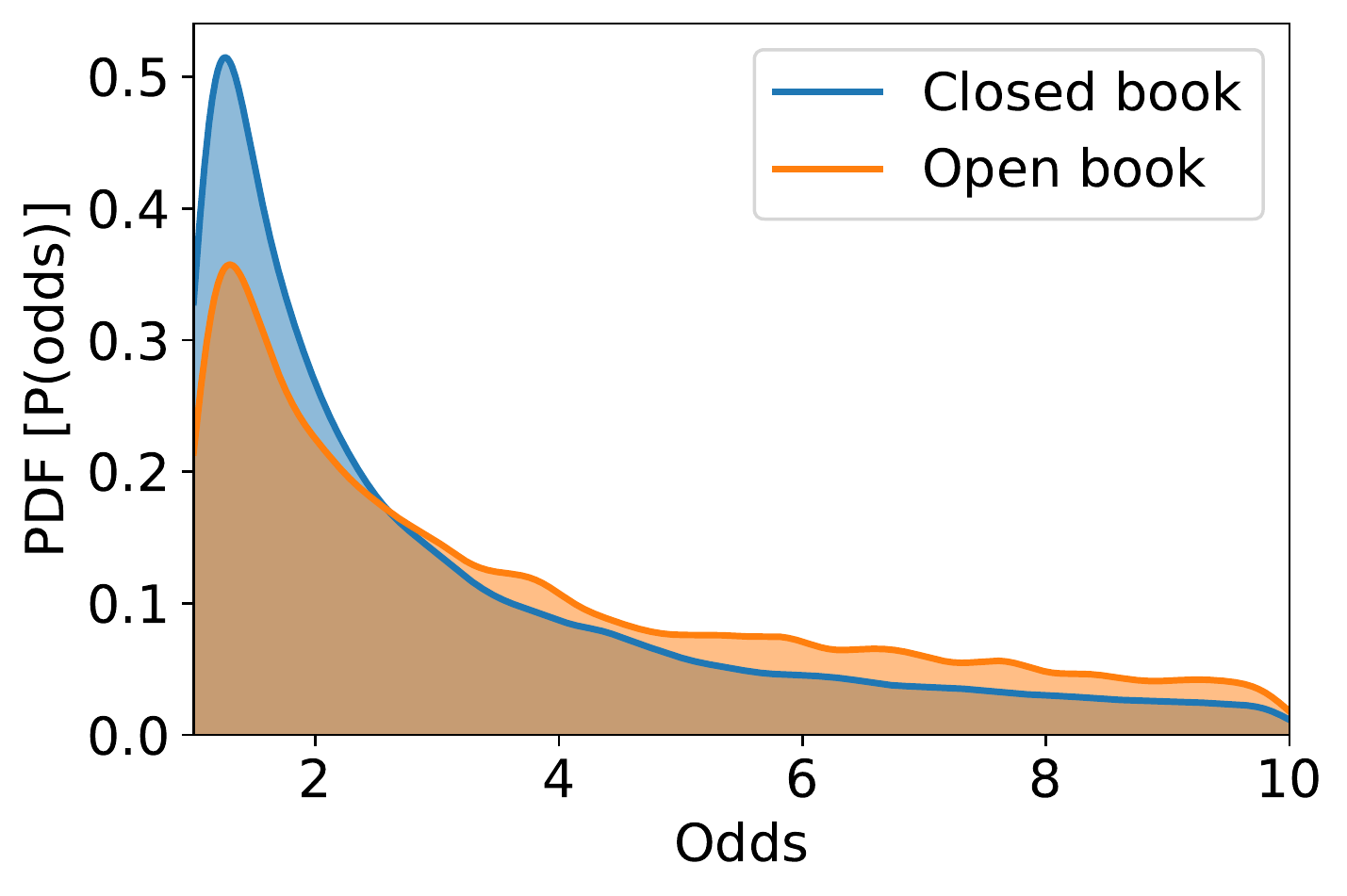}
    \caption{The figure shows the distribution of the odds between the top 2 tokens in position 1, both for the open book as well as the closed book generation. The flatter curve and fat tail for the open book response indicates that the model is more certain in its response. }
    \label{fig:odds}
\end{figure}
We first reproduce the experiment described in section~\ref{sec:robust} with the resulting data from our experiment, and the results are shown in figure~\ref{fig:odds}. We observe that when providing context to a QA task, the model demonstrates increased certainty in the answers it produces and is therefore also more robust to small changes in the prompt.

We conduct another experiment to verify this where we modify the prompt in one of two ways: 1) We perturb the dataset by introducing typos, grammatical errors, or replacing individual words with synonyms, regardless of whether the synonym changes the implicit meaning of the question or not (as can happen in English). 2) We rephrase the question. Then, we get the answers and evaluate the new answers w.r.t the original question. If Tree-Search is more robust, it should perform better than what is shown in~\ref{tab:results}.

For each alternative, we repeat the process illustrated in figure~\ref{fig:flowchart} with the altered question. The underlying idea is that the changed prompts should generate similar trees, as Tree-Search aims to sample relevant parts of the answer space effectively, regardless of the exact wording of the initial prompt. Subsequently, the model should evaluate the same best answer (to the unchanged question), as something close to it should appear in the tree even though the greedy answer often differs. We then compare the new answers using the same metrics as the previous experiment: informativeness, accuracy, coherence, and consistency. The results are displayed in table~\ref{tab:comparison}, which shows the change caused by each alternative. Also shown in the table is that for this experiment we examined the effect of increasing tree size. With an increased maximum, the difference between closed book and open book QA becomes even more significant. We interpret this as the change in question pushing the model away from the correct answer, but a larger tree allows for wider (and well-sampled) exploration of the answer space, and when the tree reaches a given size, it also samples the correct answer.

Examining the results in the table, it becomes quite apparent that there is an increase in robustness as well as performance with increased tree size. The fact that we see the largest gains in the "consistent" metric is another indication of increased robustness as this measures alignment between answer and question (but does not require it to be \textit{correct} like the accuracy metric). As for the "informative" metric, this is where the room for further gains was the smallest, which is likely why we see such a small difference. 
\begin{table}
  \centering
    \caption{The change in proportion of preferred open book answers, for two different changes to the prompt and different tree sizes.}
  \begin{tabular}{llrrr}
    \toprule
    & & \multicolumn{3}{c}{\textbf{Maximum Tree Size}} \\
    \cmidrule(lr){3-5}
    \textbf{Category} & \textbf{Q change} & \textbf{20}[\%] & \textbf{50}[\%] &  \textbf{100}[\%]\\
    \midrule
    \multirow{3}{*}{\textbf{Informative}} & Perturb & $-12.0\pm8.5$ & $-1.8\pm1.3$ & $0.8\pm0.6$ \\
    & Rephrase & $-3.2\pm2.6$ & $1.7\pm1.1$ & $9.1\pm7.1$\\
    \addlinespace
    \multirow{3}{*}{\textbf{Consistent}} & Perturb & $8.5\pm3.5$ & $15.2\pm5.9$ & $25.1\pm10.3$\\
    & Rephrase & $9.6\pm3.5$ & $8.8\pm4.3$ & $36.2\pm16.0$\\
    \addlinespace
    \multirow{3}{*}{\textbf{Accuracy}} & Perturb & $3.5\pm1.5$ & $8.7\pm4.2$ & $13.4\pm6.2$\\
    & Rephrase & $11.5\pm5.2$ & $13.0\pm6.1$ & $8.1\pm4.2$\\
    \addlinespace
    \multirow{3}{*}{\textbf{Coherence}} & Perturb & $-1.7\pm0.8$ & $4.0\pm2.1$ & $13.0\pm7.4$ \\
    & Rephrase & $7.1\pm3.7$ & $10.1\pm5.3$ & $15.2\pm8.3$\\
    \bottomrule
  \end{tabular}

  \label{tab:comparison}
\end{table}

\subsection{Tree-Search}

We introduced Tree-Search, a novel decoding strategy designed to extract diverse information from large language models. Its robustness in enhancing question-answering tasks is particularly noteworthy. However, its application extends beyond that, proving beneficial across a variety of NLP tasks, such as text summarization, machine translation, and creative text generation. The strategy is flexible, allowing for enhanced diversity and controllability of outputs, which are critical for the quality and usefulness of the results. Furthermore, by adjusting key parameters like the entropy threshold, probability cut-off, and search depth, or controlling token selection at high entropy positions, Tree-Search provides a way to steer the model towards desired responses. Therefore, while the strategy warrants further exploration of its nuances, it already strikes a balance between diversity and controllability, demonstrating its potential as a significant advancement in decoding strategies.

\subsection{Closed to open book QA}
Upon examining table~\ref{tab:results}, it becomes evident that our method of self-contextualisation successfully extracts more information from the model compared to traditional closed book prompting. This outcome demonstrates the effectiveness of Tree-Search in generating diverse outputs and highlights the benefits of providing context to the model for more informative answers, even when that additional context comes from the model itself.

Whilst the methodology often provides more informative answers, it does not show as significant improvements in other metrics. We have identified two primary reasons for this.

\noindent When utilizing Tree-Search to obtain information from the model, there is a possibility of extracting incorrect or irrelevant information. When re-prompting the model with this context, it struggles to identify and filter out incorrect details, which affects the accuracy and consistency of the generated answers. This issue highlights the limitations of the model's ability to discern the reliability of the extracted information within the context provided.

\noindent When it comes to open book QA, this type of model is primarily trained for extractive QA, i.e., processing context and extracting answers from it. Our method of constructing context consistently introduces syntactical and grammatical errors. The model often extracts the answer from this imperfect context without correcting the errors, which could negatively impact the consistency and coherence of the generated responses. 

\noindent We attempted to alleviate this issue by having the model summarise the tree, rather than using it for context in a QA; using the prompt shown below. 
\begin{tcolorbox}
\noindent \textbf{Document:} \verb|{{context}}| \\
\noindent \textbf{Summary:} 
\end{tcolorbox}
Whilst this does eliminate a large proportion of the syntactical errors from the answer, it does often lead to a misalignment between question and answer and a lower score on most of our metrics. However, this does indicate that a lot of improvement could be achieved by further refining the prompts as even with the basic context assembly and prompting we see enhanced performance.

\subsection{The Potential of Bootstrapping}
The promising results of our Tree-Search based approach open the possibility for a bootstrapping method. This technique would create a feedback loop for iterative improvement, using the enhanced outputs of the model for its retraining, potentially leading to substantial growth in the model's performance and robustness.

One area worth exploring in this context is incremental retraining. Instead of retraining the model from scratch, it could be incrementally retrained with the superior outputs yielded by our self-contextualizing QA approach. This strategy could streamline the iterative improvement process, possibly leading to quicker convergence to a more robust and high-performing model.

Further, assessing the impact of bootstrapping on model robustness would shed light on its potential for performance enhancement. This could involve evaluating the model's robustness metrics, such as out-of-distribution generalization and adversarial resilience, pre and post-bootstrapping. These investigations would offer valuable insights into the effectiveness of bootstrapping as a tool for iterative model enhancement.

\subsection{Summary and Conclusions}
This study demonstrated a notable improvement in QA tasks for smaller LLMs, specifically transforming from a closed book to an open book QA format. The  application of Tree-Search played a crucial role in this transformation, where the model's initial responses were effectively utilized to create a rich context, enabling it to augment its responses.

While the process encountered certain limitations, the quality of responses exhibited a measurable improvement in terms of accuracy, consistence, informativeness and coherence. In addition to providing improved responses it also demonstrates an increase in robustness

The transformation of the model into an open book QA system via Tree-Search certainly shows potential. Future work could focus on refining the context assembly, improving error correction in the model, and investigating alternative context selection strategies. Although promising, this approach requires further exploration and refinement to fully realize its potential in diverse domains.

\bibliography{refs.bib}
\bibliographystyle{mnras}
\newpage
\appendix

\end{document}